\documentclass{article}
\usepackage{spconf,amsmath,graphicx}
\usepackage{flushend}
\usepackage{epsfig}
\usepackage{color}
\usepackage{cite}
\usepackage{mathrsfs}
\usepackage{subfigure}
\usepackage{subfig}
\usepackage{algorithm, algpseudocode}
\usepackage{amsmath}
\usepackage{amsthm}
\usepackage{amssymb }
\usepackage{bm}
\usepackage{multirow}  
\usepackage{booktabs}
\usepackage{enumitem}
\usepackage{pbox}


\def\EPA{\text{TC-HGR}}
\def\TB{\text{Temporal Convolution Block}}
\def\TC{\text{Temporal Convolutions}}
\def\minmax{\text{Minmax}}
\def\nina{\text{Ninapro}}
\def\W{\mathit{W}}
\def\D{\mathcal{D}}
\def\X{\bm{X}}
\def\i{_{i}}
\def\y{\mathrm{y}}
\def\M{M}
\def\L{L}
\def\C{C}
\def\cl{^{\C\times\L}}
\def\P{P}
\def\d{D}
\def\r{R}

\def\Z{\bm{Z}}
\def\N{N}
\def\R{\mathbb{R}}
\def\th{^{\text{th}}}

\title{Hand Gesture Recognition Using Temporal Convolutions and Attention Mechanism}
\name{Elahe Rahimian$^{\dagger}$, Soheil Zabihi$^\ddagger$, Amir Asif$^\ddagger$, Dario Farina$^{\dagger\dagger}$,  S. Farokh Atashzar$^{\ddagger\ddagger}$, and Arash Mohammadi$^{\dagger}$}

\address{$^\dagger$Concordia Institute for Information System Engineering, Concordia University, Montreal, QC, Canada\\
$^\ddagger$Electrical and Computer Engineering,  Concordia University, Montreal, QC, Canada\\
$^{\dagger\dagger}$Department of Bioengineering, Imperial College London, London, UK\\
$^{\ddagger\ddagger}$Electrical \& Computer Engineering, Mechanical \& Aerospace Engineering, New York University, USA}

\begin{document}
\ninept
\frenchspacing
\maketitle
%
\begin{abstract}
Advances in biosignal signal processing and machine learning, in particular Deep Neural Networks (DNNs), have paved the way for the development of innovative Human-Machine Interfaces for decoding the human intent and controlling artificial limbs. DNN models have shown promising results with respect to other algorithms for decoding muscle electrical activity, especially for recognition of hand gestures. Such data-driven models, however, have been challenged by their need for a large number of trainable parameters and their structural complexity. Here we propose the novel $\TC$-based Hand Gesture Recognition architecture ($\EPA$) to reduce this computational burden. With this approach, we classified $17$ hand gestures via surface Electromyogram (sEMG) signals by the adoption of attention mechanisms and temporal convolutions. The proposed method led to $81.65\%$ and $80.72\%$ classification accuracy for window sizes of $300$ms and $200$ms, respectively. The number of parameters to train the proposed $\EPA$ architecture is $11.9$ times less than that of its state-of-the-art counterpart. 
\end{abstract}
%
%
\vspace{-.1in}
\section{Introduction} \label{intro}
\vspace{-.1in}
Hand gesture recognition via surface Electromyogram (sEMG) signals~\cite{ICASSP21_Elahe, Icassp_Elahe,Icassp, 2_Dario, Dario} has been investigated in the literature as the most promising approach for myoelectric control of prosthetic systems. In particular, Hand Gesture Recognition (HGR) has been the focus of different research works~\cite{JMRR_Elahe, Globalsip_Elahe}, given its unique potentials to improve the quality of control and consequently to enhance the quality of life of amputees. Although academic researchers have used advanced Machine Learning (ML) and Deep Neural Network (DNN) models to achieve promising laboratory results in HGR, translating these new techniques into the daily lives of amputees has faced several critical challenges~\cite{19-Patrick3, 3_Dario}. One of the key challenges is the dependency of DNN models on a large  number of trainable parameters, which leads to  structural complexity and limits their applicability to clinical settings~\cite{Atashzar}. Therefore, there is an urgent and unmet quest to develop DNN-based learning frameworks that focus on reducing the number of parameters and maintaining high performance.

The performance of DNN models developed based on sparse multi-channel sEMG is still significantly lower than that of High-Density sEMG (HD-sEMG) systems which has a high number of densely located electrodes to significantly increase the information rate~\cite{GengNet, Wei2017, HuNet}. For instance, in Reference~\cite{Wei2017}, the HGR accuracy of $99.7\%$ is reported using HD-sEMG, which is reduced to $84.4\%$ when sparse multi-channel sEMG signals are used. In this context, we aim to design a novel DNN architecture using sparse multi-channel sEMG signals provided by the $\nina$ \cite{DB1, DB1_2_3} database, which is one of the most widely accepted  sparse multi-channel sEMG benchmark datasets. We designed the novel $\TC$-Hand Gesture Recognition architecture ($\EPA$) to reduce computational burden while  maintaining high accuracy, which is of paramount importance to translate the classification results into smooth actions.

A common strategy for classifying hand movements with DNN-based algorithms is converting sEMG signals into images and then using Convolutional Neural Networks (CNNs) to detect hand movements~\cite{GengNet, Wei2017, DingNet, WeiNet}.  sEMG signals are, however, sequential in nature, and CNNs cannot extract temporal features. In this regard, recent literature~\cite{HuNet, Atashzar} used recurrent architectures such as Long Short Term Memory (LSTM) networks to consider the sequential nature of sEMG signals. In addition, LSTMs and CNNs can be combined as a hybrid architecture~\cite{JMRR_Elahe} to jointly extract the temporal and spatial properties of sEMG signals. Although sequence modeling with recurrent networks is a common approach, it can have disadvantages, such as lack of parallelism during training, exploding/vanishing gradient, and extensive memory/computation requirements~\cite{TCN}. Therefore, Reference~\cite{TCN} proposed $\TC$ (TCs) for extracting temporal information in time series tasks while addressing the aforementioned challenges of the recurrent architectures. In addition Reference~\cite{Atashzar} proposed the concept of temporal dilation of LSTM to reduce the computational cost, memory, requirement, and model complexity. On the other hand, attention-based architectures such as Transformers~\cite{Vaswani, ViT} show great potentials for widespread adoption in different Artificial Intelligence (AI) applications. In particular, the transformer-based architectures could improve the recognition accuracy compared to their state-of-the-art counterparts (where LSTM or hybrid LSTM-CNN are adopted). However, transformers are limited by the memory and computation requirements of the quadratic operation in attention for long sequences or images. Therefore, in~\cite{ViT}, the authors proposed dividing the images into patches and then using the flattened patches as the input for the transformers. Inspired by the progress of attention mechanism and TCs, recent literature~\cite{snail, TNSRE_Elahe, ICASSP21_Elahe} proposed a Few-shot learning architecture, which is based on the combination of attention and temporal convolutions.

In this paper, to address the challenges with recurrent architectures and achieve high accuracy for sparse multichannel sEMG, we proposed the design of the $\EPA$, which is based on TCs and attention mechanism. More specifically, TCs use the dilated causal convolutions, which can increase the receptive field of the network and extract the temporal features of the sEMG signals. Moreover, by inspiring from~\cite{ViT}, we divided the sEMG signals into patches to reduce the computational cost. The contributions of the paper can be summarized as follows:
\begin{itemize}[noitemsep]
\vspace{-.05in}
\item The $\EPA$ framework is proposed based on self-attention mechanism and temporal convolution to address the aforementioned challenges with the recurrent architectures.
\item The $\EPA$ reduces the number of parameters, which is a key step forward to embed the DNN models into prostheses controllers.
\item The $\EPA$ divides the sEMG signals into patches, which reduces the computational burden of the system.
\item The $\EPA$ can access a long history through temporal convolutions and also can pinpoint specific information in the sEMG signals through the attention mechanism.
\vspace{-.05in}
\end{itemize}
%

\vspace{-.1in}
\section{The Proposed $\EPA$ Architecture} \label{architecture}
\vspace{-.1in}
Developing a DNN architecture with fewer parameters can tackle the challenge of structural complexity and reduce the gap between academic research and practical settings for myoelectric prosthesis control. In what follows, first, we present the pre-processing step of the proposed $\EPA$ and then present its detailed architecture.

\vspace{-.08in}
\subsection{Preprocessing Step} \label{preprocess}
\vspace{-.05in}

\noindent
For developing $\EPA$ architecture, we used the raw time-domain sEMG signals. Following the recent literature~\cite{pattern_letter2019,GengNet, AtzoriNet,DB1_2_3}, the sEMG signals are pre-processed and smoothed using a  $1^{\text{st}}$ order low-pass Butterworth filter. Moreover, to amplify the magnitude of sensors with small values, we scaled the sEMG signals logarithmically~\cite{Icassp_Elahe}. This technique is known as \textit{$\mu$-law} transformation and is used for quantization in the speech domain. However, we use it for normalizing the sEMG signals, i.e.,
%
\begin{equation}\label{mu_law}
F(x_t) = \text{sign}(x_t)\frac{\ln{\big(1+ \mu |x_t|\big)}}{\ln{\big(1+ \mu \big)}},
\end{equation}
where $t$ represents the time; $x_t$ shows the sEMG signal, and $\mu$ is the parameter that indicates the new range. In Reference~\cite{Icassp_Elahe}, the authors showed that scaling sEMG signals with \textit{$\mu$-law} technique results in better performance than normalizing with $\minmax$. This completes the steps performed to pre-process the sEMG signals and prepare the input to be provided to the $\EPA$ architecture. Next, we present the detailed structure of the proposed architecture.

\vspace{-.08in}
\subsection{The $\EPA$ Architecture} \label{module}
\vspace{-.05in}
\setlength{\textfloatsep}{0pt}
\begin{figure}[t!]
	\centering
	\includegraphics[scale=.47]{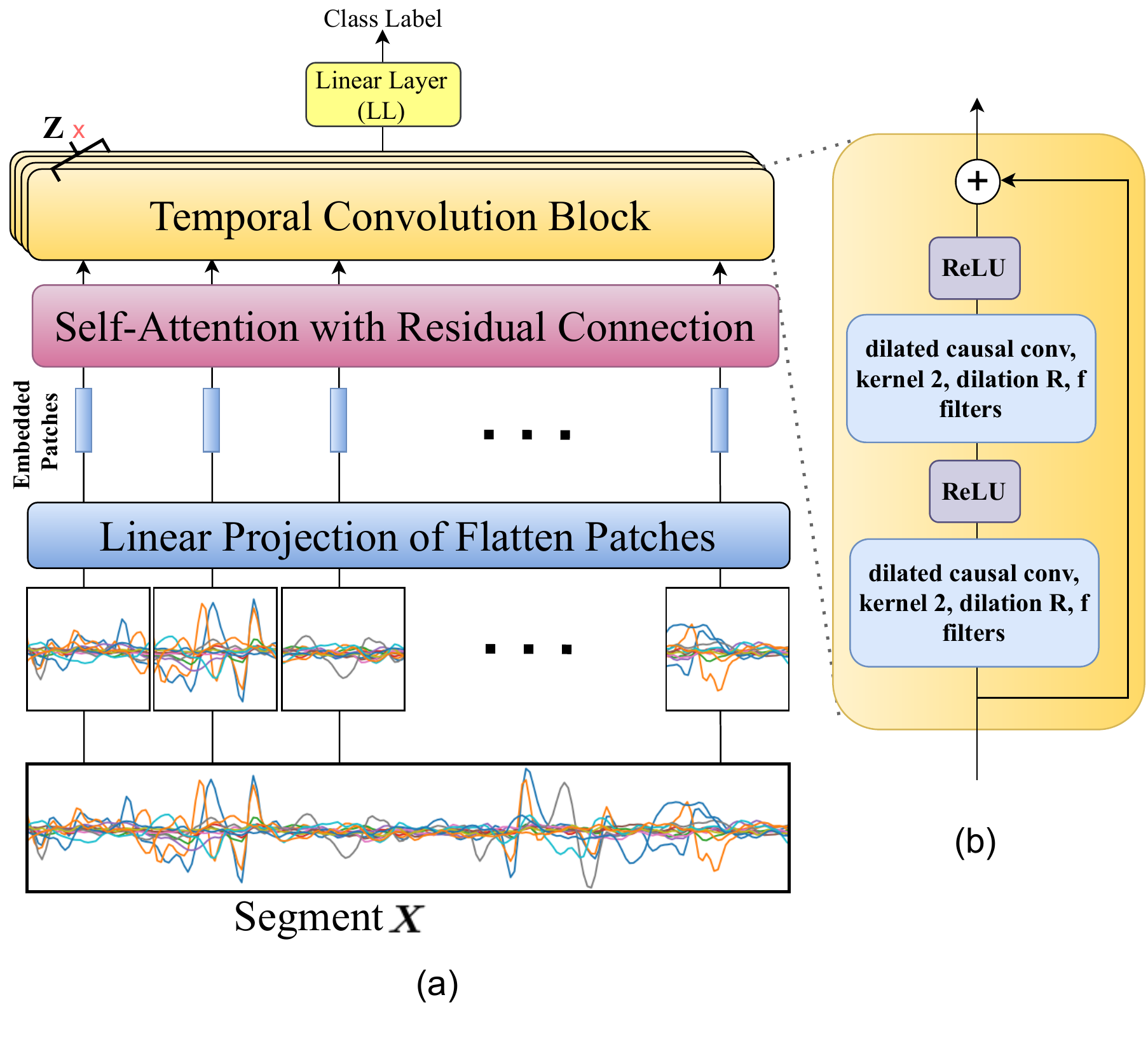}
	\vspace{-.4in}
	\caption{\footnotesize The proposed $\EPA$ architecture: (a) Each input segment $\X$ (for simplicity, we dropped the index $i$) is divided into $\N$ non-overlapping patches. Then, each patch is flattened and mapped to model dimension $\d$ (blue block). We refer to the output of this process as Embedded Patches. The sequence of the Embedded Patches is passed into the Self-Attention module, which includes the residual connection (purple block). Afterward, we used $\Z$ number of $\TB$s to access a long history (orange block). (b) Each $\TB$ consists of two dilated causal convolutions, each followed by a ReLU activation function. Again, we used residual connections to concatenate the output and input. Finally, a Linear Layer (LL) is adopted to output the class label.}
	\vspace{.1in}
	\label{arc}
\end{figure}

\noindent
In this section, the proposed $\EPA$ architecture is described in detail. After pre-processing, we segment the sEMG signals based on a window of size $\W$ $\in \{200ms, 300ms\}$, resulting in the dataset $\D = \{(\X\i, \y\i)\}_{i=1}^{\M}$. More specifically, $\X\i \in \R\cl$ is the $i^{\th}$ segment with label $\y\i$, for ($1 \leq i \leq \M$). Here, $\C$ indicates the number of channels in the input segment, and $\L$ is the length of the segmented sequence, which represents the number of samples obtained at a frequency of $2$ kHz for a window of size $\W$. As illustrated in Fig.~\ref{arc}, the $\EPA$ framework has been developed based on ``Embedded Patches'' and two modules, namely \textit{$\TB$} and \textit{Self-Attention Module with Residual Connection}, which is described below:

\noindent
(i) \textit{\textbf{Embedded Patches:}} In a similar way to the Vision Transformer (ViT) architecture~\cite{ViT}, the input segment $\X\i$ is divided into $\N$ non-overlapping patches. Here, $\N = \L / \P$, where $\P$ shows the size of each patch. This patching mechanism helps reduce memory and computation requirements. As shown in Fig.~\ref{arc}, the sequence of linear projections of these patches are fed as input to the $\EPA$ architecture. More specifically, each patch is first flattened and then is mapped into the model dimension $\d$ with a trainable linear projection. The output of this projection is called the ``Embedded Patches''. 

\noindent
(ii) \textit{\textbf{$\TB$:}} In recent literature (such as~\cite{TCN}), the authors represented TCs for the sequence modeling tasks and showed that temporal convolutions could outperform recurrent networks such as LSTMs in a wide range of datasets and time-series tasks. More specifically, TCs offer several advantages over recurrent networks such as processing the input sequence as a whole rather than sequential training, low memory requirements, stable gradient, and capturing the past information with flexible receptive field size. Inspired by the performance of TCs for the sequential data, we used ``$\TB$'' (represented via an orange box in Fig.~\ref{arc}) instead of recurrent networks for the sEMG-based HGR. As shown in Fig.~\ref{arc}, each $\TB$ consists of dilated causal convolutions, where the dilation rate $\r$ increases exponentially, i.e.,  ($\mathrm{1, 2, 4, 8, \ldots}$), to access a receptive field that exceeds the length of the input sequence.  Moreover, each dilated causal convolution is followed by a ReLU activation function. The number of $\TB$s are based on the logarithmic scale with the number of patches $\N$. More specifically, we used $Z=\left \lceil{\log_2 \N}\right \rceil$ number of $\TB$ for an input segment $\X$ with a sequence length of $\L$.

\noindent
(iii) \textit{\textbf{Self-Attention Module with Residual Connection:}} In the proposed $\EPA$ architecture, we used the ``Temporal Convolution Block'' along with the ``Attention'' mechanism. In~\cite{Vaswani}, the authors showed that the attention mechanism allows a model to present important information in a given input sequence. Moreover, the attention mechanism has recently been used~\cite{ICASSP21_Elahe} in the context of sEMG-based hand gesture recognition, where experiments have demonstrated the ability of attention to identifying specific pieces of information in the sequential nature of the sEMG signals. On the other hand, in References~\cite{snail, TNSRE_Elahe}, it was shown that temporal convolutions and attention are complementary mechanisms, i.e., the former captures a long history while the latter identifies a specific type of information.

An attention block measures the pairwise similarity of each query and all keys to assign a weight to each value. Then, the output  is computed, which is the weighted sum of the values~\cite{Vaswani}. The keys, values, and queries are packed together into matrices  $\bm{K}$, $\bm{V}$, and $\bm{Q}$, respectively. The output matrix is then computed as follows
\begin{equation}\label{attention}
\text{Attention}(\bm{Q}, \bm{K}, \bm{V}) = \text{softmax}(\frac{\bm{Q}\bm{K}^T}{\sqrt{d_k}})\bm{V},
\end{equation}
where $\mathit{d_k}$ denotes the dimension of $\bm{K}$ and $\bm{Q}$. In the $\EPA$ architecture, we also used residual connections to concatenate the output and input. This completes the description of the proposed $\EPA$ architecture, next, we present our results to evaluate its HGR performance.
\begin{table}[t!]
\centering
\renewcommand\arraystretch{1.8}
\caption{\footnotesize Descriptions of $\EPA$ architecture variants.\label{table1}}
\vspace{-.1in}
\resizebox{\columnwidth}{!}
{\begin{tabular}{  c | c | c c c c c}
\hline
\hline
\textbf{Window size $\W$}
& \textbf{Model ID}
& \textbf{Number of Patches $\N$}
& \textbf{Model dimension $\d$}
& \textbf{Params}

\\
\hline
\multicolumn{1}{c|}{\multirow{4}[2]{*}{\textbf{200ms}}}
& \textbf{1}
& 10
& 12
& 49,186
\\
& \textbf{2}
& 10
& 16
& 68,445
\\
& \textbf{3}
& 16
& 12
& 69,076
\\
& \textbf{4}
& 16
& 16
& 94,965
\\
\hline
\multicolumn{1}{c|}{\multirow{4}[2]{*}{\textbf{300ms}}}
& \textbf{1}
& 10
& 12
& 52,066
\\
& \textbf{2}
& 10
& 16
& 72,285
\\
& \textbf{3}
& 15
& 12
& 67,651
\\
& \textbf{4}
& 15
& 16
& 92,945

\\
\hline
\hline
\end{tabular}}
\end{table}
\begin{table}[t!]
\centering
\renewcommand\arraystretch{2}
\caption{\footnotesize classification accuracies for $\EPA$ architectures variants. The STD denotes the standard variation in accuracy over the $40$ subjects. \label{table2}}
\vspace{-.1in}
\resizebox{\columnwidth}{!}
{\begin{tabular}{  c c | c c c c}
\hline
\hline
\multicolumn{1}{c}{\multirow{2}[6]{*}{\rotatebox[origin=c]{90}{\footnotesize \textbf{Window size}}}} \newline \multirow{2}[6]{*}{\rotatebox[origin=c]{90}{\footnotesize \textbf{200ms}}}
& \multicolumn{1}{|c|}{\textbf{Model ID}}
& \textbf{1}
& \textbf{2}
& \textbf{3}
& \textbf{4}
\\
\cline{2-6}
&
\multicolumn{1}{|c|}{Accuracy ($\%$)} & $80.29$  & $80.63$ & $80.51$ & $\textbf{80.72}$
\\
&
\multicolumn{1}{|c|}{STD ($\%$)} & $6.7$ & $6.8$ & $6.7$ & $\textbf{6.6}$
\\
\hline
\multicolumn{1}{c}{\multirow{2}[6]{*}{\rotatebox[origin=c]{90}{\footnotesize \textbf{Window size}}}} \newline \multirow{2}[6]{*}{\rotatebox[origin=c]{90}{\footnotesize \textbf{300ms}}}
& \multicolumn{1}{|c|}{\textbf{Model ID}}
& \textbf{1}
& \textbf{2}
& \textbf{3}
& \textbf{4}
\\
\cline{2-6}
&
\multicolumn{1}{|c|}{Accuracy ($\%$)} & $80.84$ & $81.59$  & $80.95$  & $\textbf{81.65}$
\\
&
\multicolumn{1}{|c|}{STD ($\%$)} & $6.4$  & $6.5$  & $6.5$ & $\textbf{6.7}$
\\
\hline
\hline
\end{tabular}}
\vspace{.1in}
\end{table}
\begin{table*}[t!]
\centering
\renewcommand\arraystretch{2}
\caption{\small Comparison between the proposed $\EPA$ methodology and previous works~\cite{Atashzar}, which used recurrent architectures (LSTM).\label{table3}}
\vspace{-.1in}
{\begin{tabular}{c  c | c c | c c}
\hline
\hline
\multicolumn{1}{c}{\multirow{7}[2]{*}{\rotatebox[origin=c]{0}{\footnotesize \textbf{Reference~\cite{Atashzar}}}}}
&
& \multicolumn{2}{c|}{\textbf{200ms}}
& \multicolumn{2}{c}{\textbf{300ms}}
\\
\cline{3-6}
&
& \multicolumn{1}{c}{\textbf{Params}}
& \textbf{Accuracy ($\%$)}
& \multicolumn{1}{c}{\textbf{Params}}
& \textbf{Accuracy ($\%$)}
\\
\cline{1-6}
& \multicolumn{1}{|c|}{4-layer 3rd Order Dilation}
& $1,102,801$
& $79.0$
& $1,102,801$
& $82.4$
\\
& \multicolumn{1}{|c|}{4-layer 3rd Order Dilation (pure LSTM)}
& $\_$
& $\_$
& $466,944$
& $79.7$
\\
& \multicolumn{1}{|c|}{SVM}
& $\_$
& $26.9$
& $\_$
& $30.7$
\\
\hline
\multicolumn{1}{c}{\multirow{2}[2]{*}{\rotatebox[origin=c]{0}{\footnotesize \textbf{Our Method}}}}
& \multicolumn{1}{|c|}{Model 1}
& $\textbf{49,186}$
& $80.29$
& \textbf{52,066}
& $80.84$
\\
& \multicolumn{1}{|c|}{Model 4}
& $94,965$
& $\textbf{80.72}$
& $92,945$
& $\textbf{81.65}$

\\
\hline
\hline
\end{tabular}}
\vspace{.05in}
\end{table*}
\vspace{-.08in}
\section{Experiments and Results} \label{results}
\vspace{-.1in}
In this section, first, the database used for evaluation is described followed by the presentation of different experiments and results.

\vspace{-.08in}
\subsection{Database} \label{database}
\vspace{-.05in}

\noindent
To train and evaluate the proposed $\EPA$ architecture, we used the second $\nina$ dataset~\cite{DB1_2_3} called DB2, which is a widely used public dataset. More specifically, DB2  is collected by the Delsys Trigno Wireless EMG system, which has $12$ channels and records the electrical activities of muscles at $2$ kHz. Moreover, the DB2 dataset consists of the sEMG signals from $40$ healthy subjects ($28$ males and $12$ females with age $29.9 \pm 3.9$ years among which $34$ are right-handed and six are left-handed) performing $50$ different hand gestures. Each gesture is repeated $6$ times, each lasting for $5$ seconds followed by $3$ seconds of rest. The $50$ different gestures in the DB2 dataset are presented in three sets of exercises (i.e., B, C, and D). In this paper, we focus on Exercise B, which consists of $17$ different gestures. More specifically, Exercise B consists of $9$ basic wrist movements with $8$ isometric and isotonic hand movements.
Following the recommendations provided by the $\nina$ dataset, the training set consists of $2/3$ of the repetitions of each gesture (i.e., $1, 3, 4$, and $6$), and the test set consists of the remaining repetitions (i.e., $2$ and $5$). Please refer to~\cite{DB1_2_3} for a more detailed description of the $\nina$ database.

\vspace{-.08in}
\subsection{Results and Discussions} \label{Results}
\vspace{-.05in}

\noindent
In this section, the performance of the proposed $\EPA$ architecture is evaluated through a comprehensive set of experiments. In Table~\ref{table1}, different variants of the $\EPA$ architecture are presented based on a window size of $\W$ $\in \{200ms, 300ms\}$.  For training, Adam optimizer was used across all the models with a learning rate of $0.0001$. Furthermore,  we used a mini-batch size of $32$. The performance of each model is evaluated using Cross-entropy loss. In Table~\ref{table2}, the averaged recognition accuracy of the proposed $\EPA$ architecture and its variants are reported over all subjects. In what follows, we focus on three different experiments:

\noindent
\textbf{\textit{Experiment 1 - Effect of the Model's Dimension $\d$:}} In this experiment, the objective is to investigate the effect of $\d$ of the proposed $\EPA$ architecture on the recognition accuracy. In this regard, Table~\ref{table2} has shown the results for $\d$ $\in \{12, 16\}$ for both window sizes. From Table~\ref{table1} and Table~\ref{table2}, it is observed that the accuracy
of the model will improve when the $\d$ is increased from $12$ to $16$ for the same Number of Patches $\N$.
More specifically, ``Model $2$ versus Model $1$'' and ``Model $4$ versus Model $3$'' are more accurate in both the $200$ms and $300$ms window sizes. However, from Table~\ref{table1}, it can be observed that  the number of trainable parameters has increased when $\d$ is increased from $12$ to $16$, which leads to more complexity. For instance, for $\W=200$ms and $\N=10$, Model $2$ has $68,445$ parameters, while this number is $49,186$ for Model~$1$ (Table~\ref{table1}). While increasing $\d$ can potentially improve performance, 
the implementation of prosthetic controllers is limited by its structural complexity.

\noindent
\textbf{\textit{Experiment 2 - Effect of the Number of Patches $\N$:}} This experiment is included to evaluate the effect of increasing $\N$ on the performance of the proposed $\EPA$. From Tables~\ref{table1} and~\ref{table2}, it is observed that for the same $\W$ and $\d$, accuracy increases as the number of patches $\N$ increases from $10$ to $16$. More specifically, ``Model $3$ versus Model $1$'' and ``Model $4$ versus Model $2$'' classified the hand gestures with higher accuracies in both window sizes. This is because use of more patches results in a larger effective sequence length, which in turn improves the overall performance. Increasing the number of patches, however, makes the structure more complex. For instance, for $\W=300$ms and $\d=12$, Model $3$ has $67,651$ parameters, while  Model $1$ has $52,066$ parameters (Table~\ref{table1}).

\begin{figure}[t!]
	\centering     
	\includegraphics[scale = .65]{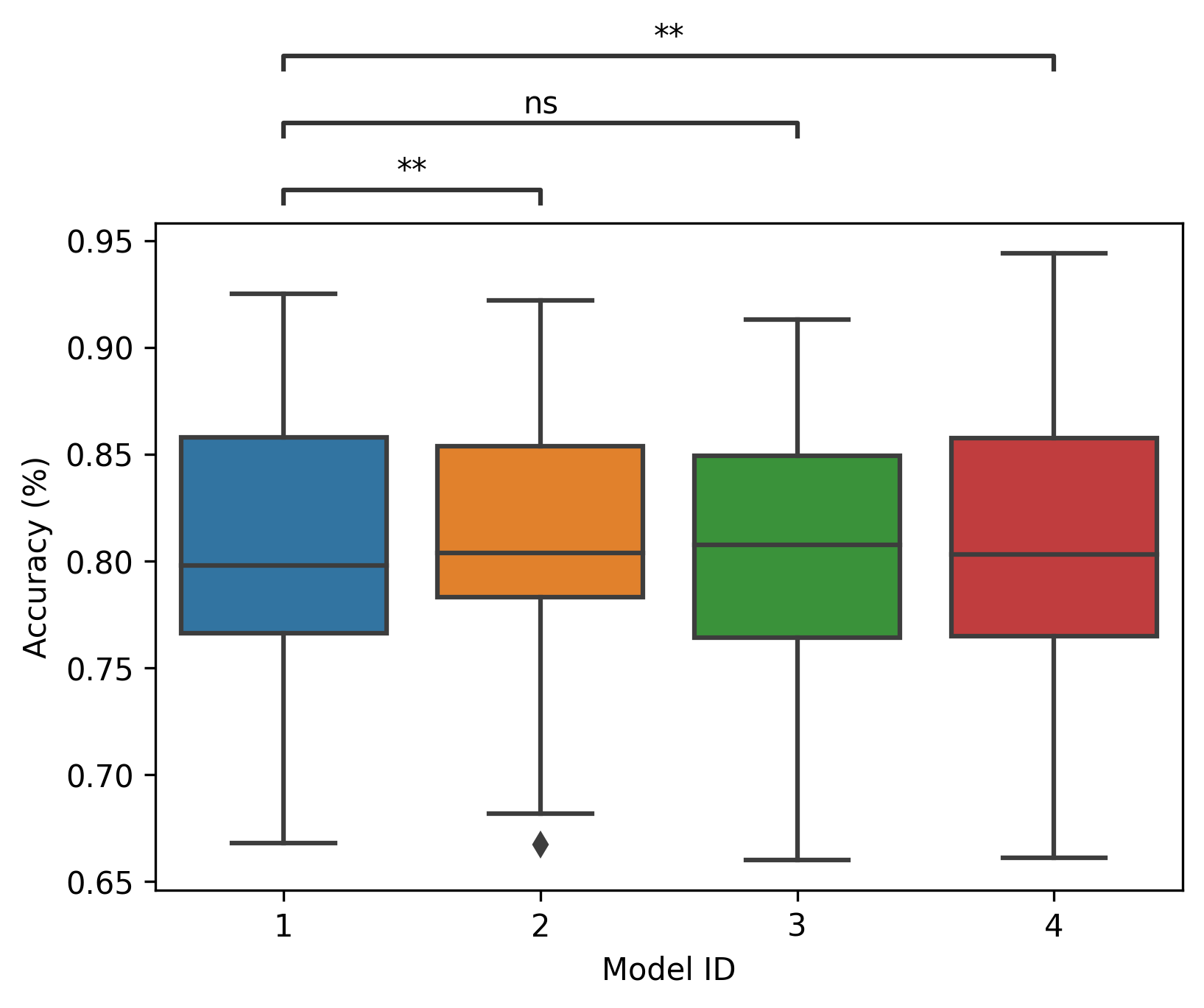}
	\caption{\footnotesize The accuracy boxplots for all $\EPA$ architecture variants for window size $300$ms. Each boxplot is representing the IQR of each model across $40$ users. We used Wilcoxon signed-rank to compare the model with the least number of paramaters (i.e.,Model $1$) with the remaining models; i.e., Model $2$, Model $3$, and Model $4$. (ns: $0.05 < \text{p-value} \leq 1$, $^{*}: 0.01 < \text{p-value} \leq 0.05$, $^{**}: 0.001 < \text{p-value} \leq 0.01$, $^{***}: 0.0001 < \text{p-value} \leq 0.001$, $^{****}: \text{p-value} \leq 0.0001$).\label{boxplot}}
	\vspace{.1in}
\end{figure}

\noindent
\textbf{\textit{Experiment 3 - Effect of Window size $\W$:}} As shown in Table~\ref{table2}, increasing the window size leads to more accuracy. This is because the larger the window size, the more information is provided for the proposed $\EPA$ architecture. In other words, machine learning model would have higher exposure to the signals from the gesture. For instance, for both ``Models $2$  and $1$'', the window size $300$ms leads to greater accuracy and complexity. However, both ``Models $3$ and $4$'' classified hand gestures more accurately while reducing the complexity because number of patches for $\W=300ms$ is less than $\W=200ms$. Although choosing a larger window size increases accuracy, using shorter windows (e.g., $200$ms) provides extra time ($100$ms)  for pre-processing or post-processing tasks.

\noindent
\textbf{\textit{Statistical Comparisons of the Different $\EPA$ Variants for Window Size 300ms:}} To examine the importance and significance of $\EPA$ variants, we perform statistical tests for all models considering $\W=300ms$. By following~\cite{TNSRE_Elahe}, the Wilcoxon signed-rank test~\cite{Wilcoxon} is used in which each user is considered as a separate dataset. As illustrated in Fig.~\ref{boxplot}, we conduct comparison between the model with the least number of parameters (i.e., Model $1$) and other models. (i.e., Model $2$, $3$, and $4$). Using Wilcoxon signed-rank test, it can be mentioned that no statistical significance was observed comparing Model $1$ and Model $3$ in Fig.~\ref{boxplot}. However, 
Model $1$ is  significant versus Model $2$ (or Model $4$) because the $p$-value is less than $0.05$. Further details are provided in Fig.~\ref{boxplot} where the $p$-value is indicated by the following symbols:
\begin{itemize}
\item Not significant (ns): p-value is between 0.05 and 1,
\item $^{*}:$ p-value is between 0.01 and 0.05,
\item $^{**}:$ p-value is between 0.001 and 0.01,
\item $^{***}:$ p-value is between 0.0001 and 0.001,
\item $^{****}:$p-value is smaller than 0.0001.
\end{itemize}
The performance distribution between $40$ users for each model is illustrated in Fig.~\ref{boxplot}. More specifically, each model is represented by a boxplot in which the performance distribution is divided into quartiles across all users (i.e., Interquartile Range (IQR)), and the median performance is represented by a horizontal line.

\noindent
\textbf{\textit{Comparison with the State-of-the-art Research~\cite{Atashzar}:}}
We have also compared the results with a recent state-of-the-art model~\cite{Atashzar} in which the same dataset is used for performance evaluations. More specifically, Reference~\cite{Atashzar} proposed models based on recurrent architectures (i.e., LSTM) with dilation. As shown in Table~\ref{table3}, for window size of $200$ms, our methodology can outperform both recurrent networks and traditional ML approaches such as Support Vector Machine (SVM). For instance, it can be observed that the accuracy for our proposed Model $4$ is $80.72\%$ with only $94,965$ number of parameters, while the best accuracy for the Reference~\cite{Atashzar} is $79\%$ with $1,102,801$ parameters.
Moreover, as shown in Table~\ref{table3}, for window size of $300$ms, the accuracy of the proposed Model $4$ is $81.65\%$, while for ``pure LSTM with dilation'' and SVM proposed in~\cite{Atashzar}, the accuracy is $79.9\%$ and $30.7\%$, respectively. Although in~\cite{Atashzar}  the authors reached to $82.4\%$  accuracy with ``4-layer 3rd Order Dilation'' which is a hybrid dilation-based LSTM, they used $1,102,801$ number of parameters which is $11.9$ times larger than the number of parameters used in our proposed Model $4$. Therefore, we provided a compact DNN model with a far fewer number of trainable parameters compared to previous works, and took a step forward towards designing more proportional, intuitive, and dexterous prostheses for clinical applications. 


\vspace{-.1in}
\section{Conclusion}\label{sec:page}
\vspace{-.08in}
In this paper, we proposed a novel architecture referred to as the $\EPA$ for Hand Gesture Recognition from sparse multichannel sEMG signals. The proposed model showed strong capability in addressing several existing challenges of gesture recognition based on the temporal convolutions and attention mechanism. We showed that by proper design of convolution-based architectures, we can extract temporal information of the sEMG signal and improve the performance. Moreover, the proposed architecture can reduce the required number of trainable parameters with respect to the state-of-the-art, which is a key enabling factor to reduce the complexity and embed DNN-based models into prostheses controllers. 



\begin{thebibliography}{10}
\footnotesize

\bibitem{ICASSP21_Elahe}
E. Rahimian, S. Zabihi, A. Asif, S.F. Atashzar, and A. Mohammadi,
\newblock ``Few-Shot Learning for Decoding Surface Electromyography for Hand Gesture Recognition,''
\newblock {\em IEEE International Conference on Acoustics, Speech and Signal Processing (ICASSP)}, 2021, pp. 1300-1304.

\bibitem{Icassp_Elahe}
E. Rahimian, S. Zabihi, F. Atashzar, A. Asif, A. Mohammadi,
\newblock ``XceptionTime: Independent Time-Window XceptionTime Architecture for Hand Gesture Classification,''
\newblock {\em International Conference on Acoustics, Speech, and Signal Processing (ICASSP)}, pp. 1304-1308, 2020.

\bibitem{Icassp}
P. Tsinganos, B. Cornelis, J. Cornelis, B, Jansen, and A. Skodras,
\newblock ``Improved Gesture Recognition Based on sEMG Signals and TCN,''
\newblock {\em International Conference on Acoustics, Speech, and Signal Processing (ICASSP)}, pp. 1169-1173, 2019.

\bibitem{2_Dario}
N. Jiang, S. Dosen, K.R. Muller, D. Farina,
\newblock ``Myoelectric Control of Artificial Limbs. Is There A Need to Change Focus?''
\newblock {\em IEEE Signal Processing Magazine}, vol. 29, pp. 150-152, 2012.

\bibitem{Dario}
D. Farina, R. Merletti, R.M. Enoka,
\newblock ``The Extraction of Neural Strategies from the Surface EMG,''
\newblock {\em Journal of Applied Physiology}, vol. 96, pp. 1486-95, 2004.

\bibitem{JMRR_Elahe}
E. Rahimian, S. Zabihi, S. F. Atashzar, A. Asif, and A. Mohammadi,
\newblock ``Surface EMG-Based Hand Gesture Recognition via Hybrid and Dilated Deep Neural Network Architectures for Neurorobotic Prostheses,''
\newblock {\em Journal of Medical Robotics Research}, pp. 1-12, 2020.

\bibitem{Globalsip_Elahe}
E. Rahimian, S. Zabihi, S. F. Atashzar, A. Asif, and A. Mohammadi,
\newblock ``Semg-based Hand Gesture Recognition via Dilated Convolutional Neural Networks,''
\newblock {\em Global Conference on Signal and Information Processing, GlobalSIP}, 2019.

\bibitem{19-Patrick3}
C. Castellini, \textit{et al.},
\newblock ``Proceedings of the First Workshop on Peripheral Machine Interfaces: Going Beyond Traditional Surface Electromyography,''
\newblock {\em Frontiers in neurorobotics}, 8, p. 22, 2014.

\bibitem{3_Dario}
D. Farina, \textit{et al.},
\newblock ``The Extraction of Neural Information from the Surface EMG for the Control of Upper-limb Prostheses: Emerging Avenues and Challenges,''
\newblock {\em Transactions on Neural Systems and Rehabilitation Engineering},  vol 22, no.4, pp. 797-809, 2014.

\bibitem{Atashzar}
T. Sun, Q. Hu, P. Gulati, and S.F. Atashzar,
\newblock ``Temporal Dilation of Deep LSTM for Agile Decoding of sEMG: Application in Prediction of Upper-limb Motor Intention in NeuroRobotics.,''
\newblock {\em IEEE Robotics and Automation Letters},  2021.

\bibitem{GengNet}
W. Geng, \textit{et al.},
\newblock ``Gesture Recognition by Instantaneous Surface EMG Images,''
\newblock {\em Scientific reports}, 6, p. 36571,  2016.

\bibitem{Wei2017}
W. Wei, Y.Wong, Y. Du, Y. Hu, M. Kankanhalli, and  W. Geng,
\newblock ``A multi-stream convolutional neural network for sEMG-based gesture recognition in muscle-computer interface.,''
\newblock {\em Pattern Recognition Letters}, 2017.


\bibitem{HuNet}
Y. Hu, \textit{et al.},
\newblock ``A Novel Attention-based Hybrid CNN-RNN Architecture for sEMG-based Gesture Recognition,''
\newblock {\em PloS one}, vol. 13, no. 10, p.e0206049, 2018.

\bibitem{DB1}
M. Atzori, A. Gijsberts, I. Kuzborskij, S. Heynen, A.G.M Hager, O. Deriaz, C. Castellini, H. Müller, and B. Caputo,
\newblock ``A Benchmark Database for Myoelectric Movement Classification,''
\newblock {\em Transactions on Neural Systems and Rehabilitation Engineering}, 2013.

\bibitem{DB1_2_3}
M. Atzori, \textit{et al.},
\newblock ``Electromyography Data for Non-Invasive Naturally-Controlled Robotic Hand Prostheses,''
\newblock {\em Scientific data 1}, vol. 1, no. 1, pp. 1-13,  2014.

\bibitem{DingNet}
Z. Ding, \textit{et al.},
\newblock ``sEMG-based Gesture Recognition with Convolution Neural Networks,''
\newblock {\em Sustainability 10}, no. 6, p. 1865,  2018.

\bibitem{WeiNet}
W. Wei, \textit{et al.},
\newblock ``Surface Electromyography-based Gesture Recognition by Multi-view Deep Learning,''
\newblock {\em IEEE Transactions on Biomedical Engineering}, vol. 66, no. 10, pp. 2964-2973, 2019.

\bibitem{TCN}
S. Bai, J.Z. Kolter, and V. Koltun,
\newblock `` An Empirical Evaluation of Generic Convolutional and Recurrent Networks for Sequence Modeling,''
\newblock {\em arXiv preprint arXiv:1803.01271}, 2018.

\bibitem{ViT}
A. Dosovitskiy, et al.,
\newblock ``An Image is Worth 16x16 Words: Transformers for Image Recognition at Scale,''
\newblock {\em  arXiv preprint arXiv:2010.11929}, 2020.

\bibitem{Vaswani}
A. Vaswani, N. Shazeer, J. Uszkoreit, L. Jones, A. Gomez N., L. Kaiser, and
I. Polosukhin,
\newblock ``Attention is All You Need,''
\newblock {\em  arXiv preprint arXiv:1706.03762}, 2017a.

\bibitem{snail}
N. Mishra, M. Rohaninejad, X. Chen, and P. Abbeel,
\newblock ``A Simple Neural Attentive Meta-Learner,''
\newblock {\em arXiv preprint arXiv:1707.03141}, 2017.


\bibitem{TNSRE_Elahe}
E. Rahimian, S. Zabihi, A. Asif, D. Farina, S.F. Atashzar, and A. Mohammadi,
\newblock ``FS-HGR: Few-shot Learning for Hand Gesture Recognition via ElectroMyography,''
\newblock {\em IEEE Trans. Neural Syst. Rehabil. Eng.}, 2021.





\bibitem{pattern_letter2019}
W. Wei, \textit{et al.}
\newblock ``A Multi-stream Convolutional Neural Network for sEMG-based Gesture Recognition in Muscle-computer Interface,''
\newblock {\em Pattern Recognition Letters}, 119, pp. 131-138,  2019.




\bibitem{AtzoriNet}
M. Atzori, M. Cognolato, and H. Müller,
\newblock ``Deep Learning with Convolutional Neural Networks Applied to Electromyography Data: A Resource for the Classification of Movements for Prosthetic Hands,''
\newblock {\em Frontiers in neurorobotics 10}, p.9, 2016.

\bibitem{Wilcoxon}
F. Wilcoxon,
\newblock ``Individual comparisons by ranking methods,''
\newblock {\em Biometrics Bull.},vol. 1, no. 6, pp. 80–83, 1945.







\end{thebibliography}
\end{document}